\def\year{2020}\relax
\documentclass[letterpaper]{article} %
\usepackage{aaai20}  %
\usepackage{times}  %
\usepackage{helvet} %
\usepackage{courier}  %
\usepackage[hyphens]{url}  %
\usepackage{graphicx} %
\usepackage{graphicx}  %
\usepackage{subcaption}
\usepackage{amsmath}
\usepackage{makecell}
\usepackage{multirow}
\usepackage{amssymb}
\usepackage[colorlinks, citecolor=black]{hyperref}
\title{Training-Time-Friendly Network for Real-Time Object Detection\thanks{Deng Cai is the corresponding author}}
\author{
Zili Liu,\textsuperscript{\rm 1,2}
Tu Zheng,\textsuperscript{\rm 1,2}
Guodong Xu,\textsuperscript{\rm 1,2}
Zheng Yang,\textsuperscript{\rm 2}
Haifeng Liu,\textsuperscript{\rm 1}
Deng Cai\textsuperscript{\rm 1,2,3} \\ %
\textsuperscript{\rm 1}State Key Lab of CAD\&CG, Zhejiang University, Hangzhou, China \\ 
\textsuperscript{\rm 2}Fabu Inc., Hangzhou, China \\
\textsuperscript{\rm 3}Alibaba-Zhejiang University Joint Institute of Frontier Technologies, Hangzhou, China\\  %
\{zililiuzju, zhengtuzju\}@gmail.com \quad yangzheng@fabu.ai \quad \{memoiry, haifengliu, dcai\}@zju.edu.cn %
}
 
\begin{document}

\maketitle

\begin{abstract}
Modern object detectors can rarely achieve short training time, fast inference speed, and high accuracy at the same time. To strike a balance among them, we propose the Training-Time-Friendly Network (TTFNet). In this work, we start with light-head, single-stage, and anchor-free designs, which enable fast inference speed. Then, we focus on shortening training time. We notice that encoding more training samples from annotated boxes plays a similar role as increasing batch size, which helps enlarge the learning rate and accelerate the training process. To this end, we introduce a novel approach using Gaussian kernels to encode training samples. Besides, we design the initiative sample weights for better information utilization. Experiments on MS COCO show that our TTFNet has great advantages in balancing training time, inference speed, and accuracy. It has reduced training time by more than seven times compared to previous real-time detectors while maintaining state-of-the-art performances. In addition, our super-fast version of TTFNet-18 and TTFNet-53 can outperform SSD300 and YOLOv3 by less than one-tenth of their training time, respectively. The code has been made available at \url{https://github.com/ZJULearning/ttfnet}.
\end{abstract}

\section{Introduction}

Accuracy, inference speed, and training time of object detectors have been widely concerned and continuously improved. However, little work can strike a good balance among them. Intuitively, detectors with faster inference speed should have a shorter training time. Nevertheless, in fact, most real-time detectors require longer training time than non-real-time ones. The high-accuracy detectors can be roughly classified into one of the two types --- those suffer from slow inference speed, and those require a large amount of training time.

The first type of networks \cite{ren2015faster,lin2017focal,DBLP:journals/corr/abs-1904-01355} generally rely on the heavy detection head or complex post-processing. Although these designs are beneficial for accuracy improvement and fast convergence, they significantly slow down the inference speed. Therefore, this type of network is typically not suitable for real-time applications.

To speed up the inference, researchers strive to simplify the detection head and post-processing while retaining accuracy \cite{liu2016ssd,redmon2018yolov3}. In a recent study named CenterNet \cite{DBLP:journals/corr/abs-1904-07850}, the inference time is further shortened --- almost the same as the time consumed by the backbone network. However, all these networks inevitably require long training time. This is because these networks are difficult to train due to the simplification, making them heavily dependent on the data augmentation and long training schedule. For example, CenterNet needs 140-epochs training on public dataset MS COCO \cite{lin2014microsoft}. In contrast, the first type of network usually requires 12 epochs.

In this work, we focus on shortening the training time while retaining state-of-the-art real-time detection performances. Previous study \cite{DBLP:journals/corr/GoyalDGNWKTJH17} has reported that a larger learning rate can be adopted if the batch size is larger, and they follow a linear relationship under most conditions. We notice that encoding more training samples from annotated boxes is similar to increasing the batch size. Since the time spent on encoding features and calculating losses is negligible compared with that of feature extraction, we can safely attain faster convergence basically without additional overhead. In contrast, CenterNet, which merely focuses on the object center for size regression, loses the opportunity to utilize the information near the object center. This design is confirmed to be the main reason for the slow convergence according to our experiments.

To shorten the training time, we propose a novel approach using Gaussian kernels to encode training samples for both localization and regression. It allows the network to make better use of the annotated boxes to produce more supervised signals, which provides the prerequisite for faster convergence. Specifically, a sub-area around the object center is constructed via the kernel, and then dense training samples are extracted from this area. Besides, the Gaussian probabilities are treated as the weights of the regression samples to emphasize those samples near the object center. We further apply appropriate normalization to take advantage of more information provided by large boxes and retain the information given by small boxes. Our approach can reduce the ambiguous and low-quality samples without requiring any other components, e.g., Feature Pyramid Network (FPN) \cite{lin2017feature}. Moreover, it does not require any offset predictions to aid in correcting the results, which is effective, unified, and intuitive.

Together with the light-head, single-stage, and anchor-free designs, this paper presents the first object detector that achieves a good balance among training time, inference speed, and accuracy. Our TTFNet reduces training time by more than seven times compared to CenterNet and other popular real-time detectors while retaining state-of-the-art performances. Besides, the super-fast version of TTFNet-18 and TTFNet-53 can achieve 25.9 AP / 112 FPS only after 1.8 hours and 32.9 AP / 55 FPS after 3.1 hours of training on 8 GTX 1080Ti, which is the shortest training time to reach these performances on MS COCO currently as far as we know. Furthermore, TTFNet-18 and TTFNet-53 can achieve 30.3 AP / 113 FPS after 19 hours and 36.2 AP / 55 FPS after 32 hours when training from scratch, and the long-training version of TTFNet-53 can achieve 39.3 AP / 57 FPS after 31 hours of training. These performances are very competitive compared to any other state-of-the-art object detectors.

Our contributions can be summarized as follows:
\begin{itemize}
\item We discuss and validate the similarity between the batch size and the number of encoded samples produced by annotated boxes. Further, we experimentally verify the main reason for the slow convergence of advanced real-time detector CenterNet.
\item We propose a novel and unified approach which uses Gaussian kernels to produce training samples for both center localization and size regression in anchor-free detectors. It shows great advantages over previous designs.
\item Without bells and whistles, our detector has reduced training time by more than seven times compared to previous real-time detectors while keeping state-of-the-art real-time detection performance. Besides, the performances of from-scratch-training and long-training version are also very significant.
\item The proposed detector is friendly to researchers, especially for who only have limited computing resources. Besides, it is suitable for training time-sensitive tasks such as Neural Architecture Search (NAS).
\end{itemize}

\section{Related Works}

\paragraph{Single Stage Detectors.} YOLO \cite{redmon2016you} and SSD \cite{liu2016ssd} achieve satisfying performances and make the single-stage network gain attention. Since focal loss is proposed \cite{lin2017focal} to solve the imbalance between positive and negative examples, single-stage detectors are considered promising to achieve similar accuracy as two-stage ones. However, after that, the accuracy of single-stage ones stagnates for a long time until CornerNet \cite{law2018cornernet} is introduced. CornerNet is a keypoint-based single-stage detector, which outperforms a range of two-stage detectors in accuracy. Its design opens a new door for the object detection task.

\paragraph{Anchor Free Design.} DenseBox \cite{DBLP:journals/corr/HuangYDY15} is the first anchor-free detector, and then UnitBox \cite{yu2016unitbox} upgrades DenseBox for better performance. YOLO is the first successful universal anchor-free detector. However, anchor-based methods \cite{ren2015faster,liu2016ssd} can achieve higher recalls, which offers more potential for performance improvement. Thus, YOLOv2 \cite{redmon2017yolo9000} abandons the previous anchor-free design and adopts the anchor design. Yet, CornerNet brings the anchor-free designs back into spotlight. Recently proposed CenterNet \cite{DBLP:journals/corr/abs-1904-08189} reduces the false detection in CornerNet, which further improves the accuracy. Apart from corner-based anchor-free design, many anchor-free detectors relying on FPN are proposed such as FCOS \cite{DBLP:journals/corr/abs-1904-01355} and FoveaBox \cite{DBLP:journals/corr/abs-1904-03797}. GARPN \cite{wang2019region} and FSAF \cite{DBLP:journals/corr/abs-1903-00621} also adopt the anchor-free design in their methods. On the contrary, CenterNet \cite{DBLP:journals/corr/abs-1904-07850} does not rely on complicated decoding strategies or heavy head designs, which can outperform popular real-time detectors \cite{liu2016ssd,redmon2018yolov3} while having faster inference speed.

\section{Motivation}

We notice that encoding more training samples plays a similar role as increasing the batch size, and both of them can provide more supervised signals for each training step. The training samples refer to the features encoded by the annotated box. Reviewing the formulation of Stochastic Gradient Descent (SGD), the weight updating expression can be described as:

\begin{equation} \label{sgd}
\begin{aligned}
w_{t+1}=w_{t}-\eta \frac{1}{n}\sum_{x\in B}\bigtriangledown l(x, w_t)
\end{aligned}
\end{equation}

\noindent where $w$ is the weight of the network, $B$ is a mini-batch sampled from the training set, $n=|B|$ is the mini-batch size, $\eta$ is the learning rate and $l(x, w)$ is the loss computed from the labeled image $x$.

As for object detection, the image $x$ may incorporate multiple annotated boxes, and these boxes will be encoded to training sample $s\in S_x$. $m_x=|S_x|$ indicates the number of samples produced by all the boxes in image $x$. Therefore (\ref{sgd}) can be formulated as:

\begin{equation} \label{sgd_od}
\begin{aligned}
w_{t+1}=w_{t}-\eta \frac{1}{n}\sum_{x\in B} \frac{1}{m_x}\sum_{s\in S_x}\bigtriangledown l(s, w_t)
\end{aligned}
\end{equation}

For simplify, suppose $m_x$ is same for each image $x$ in a mini-batch $B$. Focusing on the individual training sample $s$, (\ref{sgd_od}) can be rewritten as:

\begin{equation} \label{sgd_sim}
\begin{aligned}
w_{t+1}=w_{t}-\eta \frac{1}{nm}\sum_{s\in B}\bigtriangledown l(s, w_t)
\end{aligned}
\end{equation}

Linear Scaling Rule is empirically found in \cite{DBLP:journals/corr/GoyalDGNWKTJH17}. It claims that the learning rate should be multiplied by $k$ if the batch size is multiplied by $k$, unless the network is rapidly changing, or very large mini-batch is adopted. Namely, executing $k$ iterations with small mini-batches $B_j$ and learning rate $\eta$ is basically equivalent to executing $1$ iteration with large mini-batches $\cup_{j\in [0, k)} B_j$ and learning rate $k\eta$, only if we can assume $\bigtriangledown l(x,w_t) \approx \bigtriangledown l(x,w_{t+j})$ for $j<k$. This condition is usually met under large-scale, real-world data.

Instead of focusing on the labeled images $x$ as in \cite{DBLP:journals/corr/GoyalDGNWKTJH17}, we focus on the training samples $s$ here. The mini-batch size can be treated as $|B|=nm$ according to (\ref{sgd_sim}). Although the encoded samples $s\in S_x$ have a strong correlation, they are still able to contribute information with differences. We can qualitatively draw a similar conclusion: when the number of encoded training samples in each mini-batch is multiplied by $k$, multiply the learning rate by $l$, where $1\le l\le k$. 

\begin{figure}[!t]
\centering
\includegraphics[width=0.95\columnwidth]{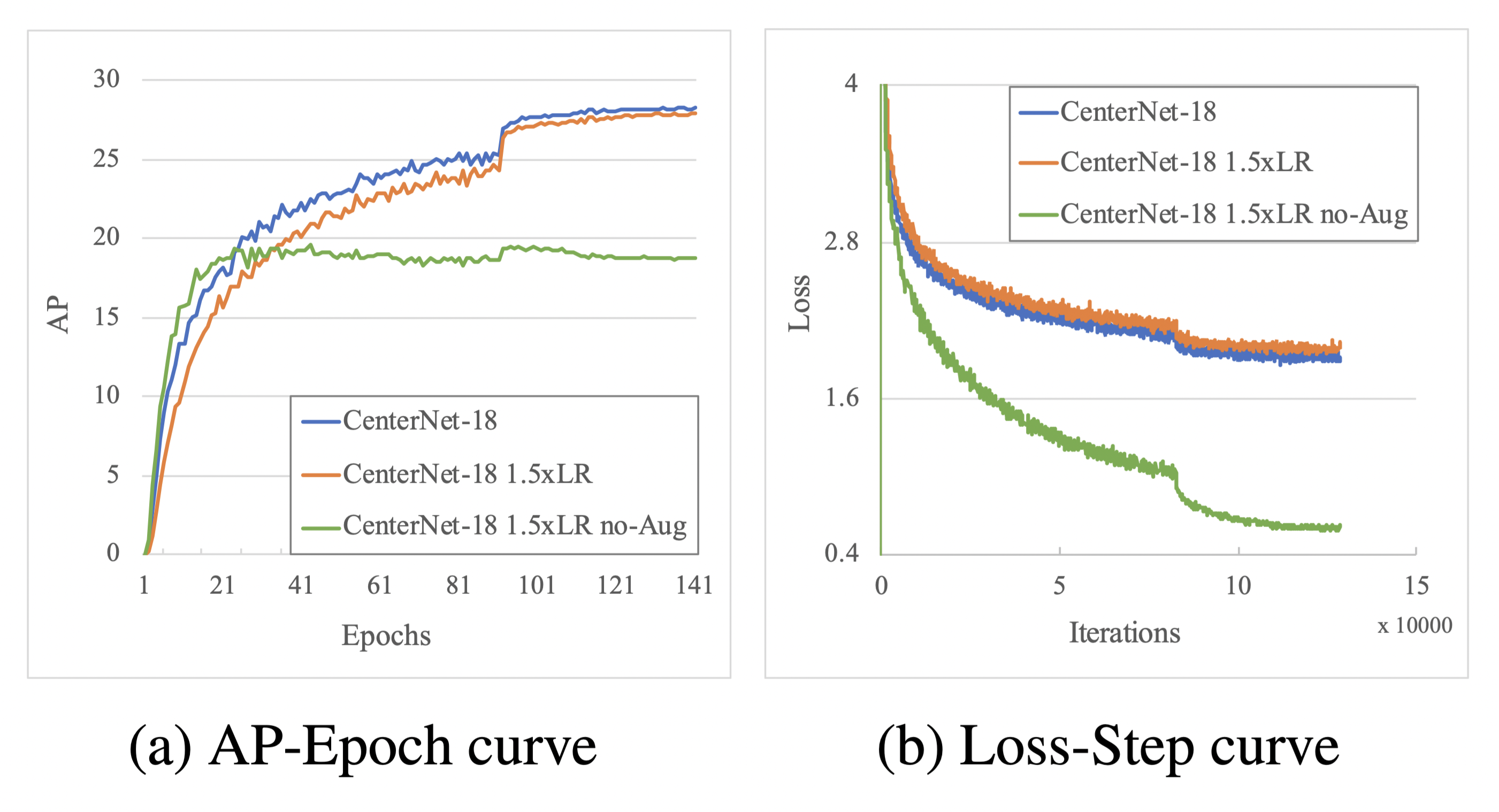}
\caption{Experiments on CenterNet-R18. We increase the learning rate by 1.5x and then remove the complex data augmentation. (a) Increasing learning rate will lead to a consistent decline in AP while (b) eliminating data augmentation will lead to obvious overfitting.}
\label{centernet}
\end{figure}

CenterNet \cite{DBLP:journals/corr/abs-1904-07850}, which is several times faster than most detectors in inference, suffers from long training time. It uses complex data augmentations in training. Although the augmentations allow models to have stable accuracy improvements, they cause slow convergence. To rule out their impact on convergence speed, we increase the learning rate and remove the augmentations. As shown in Figure \ref{centernet}, the larger learning rate cannot help CenterNet converge faster, and removing the augmentations leads to a bad performance. According to our conclusion above, we believe that it is because CenterNet merely encodes a single regression sample at the object center during training. This design makes CenterNet heavily rely on the data augmentations and long training schedule, leading to unfriendly training time.

To reduce the network's dependence on data augmentation while reducing training time, we presume that a better strategy of encoding regression samples is needed. Under the guidance of this motive, we propose our approach in the next section. More comprehensive experiments in our ablation study can further validate the superiority of our approach.

\section{Our Approach}
\subsection{Background}

CenterNet treats object detection as consisting of two parts --- center localization and size regression. For localization, it adopts the Gaussian kernel as in CornerNet to produce a heat-map, which enables the network to produce higher activations near the object center. For regression, it defines the pixel at the object center as a training sample and directly predicts the height and width of the object. It also predicts the offset to recover the discretization error caused by the output stride. Since the network can produce higher activations near the object center in inference, the time-consuming NMS can be replaced by other components with negligible overhead.

In order to eliminate the need for the NMS, we adopt a similar strategy for center localization. Specifically, we further consider the aspect ratio of the box in the Gaussian kernel since the strategy that does not consider it in CenterNet is obviously sub-optimal. 

\begin{figure}[!t]
  \centering
  \includegraphics[width=.95\columnwidth]{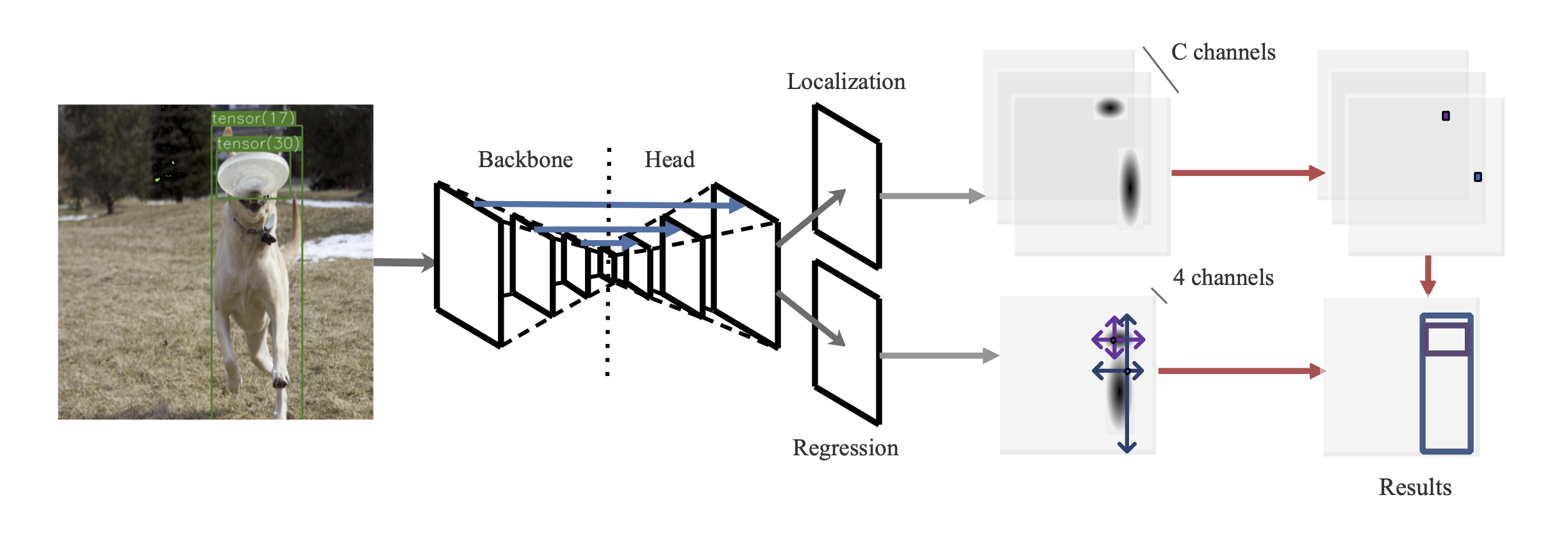}
\caption{Architecture and Pipeline of TTFNet. Features are extracted by a backbone network and then up-sampled to 1/4 resolution of the original image. Then, the features are used for localization and regression tasks. For localization, the network can produce higher activations near the object center. For regression, all samples inside the Gaussian-area of the object can directly predict the distance to four sides of the box.}
  \label{archi}
\end{figure}

As for size regression, mainstream approaches treat pixels in the whole box \cite{DBLP:journals/corr/abs-1904-01355} or the sub-rectangle area of the box \cite{DBLP:journals/corr/abs-1904-03797} as training samples. We propose to treat all pixels in a Gaussian-area as training samples. Besides, weights calculated by object size and Gaussian probability are applied to these samples for better information utilization. Note that our approach does not require any other predictions to help correct the error, as shown in Figure \ref{archi}, which is more simple and effective.

\subsection{Gaussian Kernels for Training}
Given an image, our network separately predicts feature $\hat H\in R^{N\times C\times \frac{H}{r}\times \frac{W}{r}}$ and $\hat S\in R^{N\times 4\times \frac{H}{r}\times \frac{W}{r}}$. The former is used to indicate where the object center may exist, and the latter is used to attain the information related to the object size. $N$, $C$, $H$, $W$, $r$ are batch size, number of categories, the height and width of the input image, and output stride. We set $C=80$ and $r=4$ in our experiments, and we omit $N$ later for simplify. Gaussian kernels are used in both localization and regression in our approach, and we define scalar $\alpha$ and $\beta$ to control the kernel size, respectively.

\paragraph{Object Localization.} Given $m$-th annotated box belongs to $c_m$-th category, firstly it is linearly mapped to the feature-map scale. Then, 2D Gaussian kernel K$_m(x,y)=\text{exp}(-\frac{(x-x_0)^2}{2\sigma_x^2}-\frac{(y-y_0)^2}{2\sigma_y^2})$ is adopted to produce $H_m\in R^{1\times \frac{H}{r}\times \frac{W}{r}}$, where $\sigma_x=\frac{\alpha w}{6}$, $\sigma_y=\frac{\alpha h}{6}$. Finally, we update $c_m$-th channel in $H$ by applying element-wise maximum with $H_m$. The produced $H_m$ is decided by the parameter $\alpha$, center location $(x_0, y_0)_m$, and box size $(h, w)_m$. We use ($\lfloor \frac{x}{r} \rfloor, \lfloor \frac{y}{r}\rfloor$) to force the center to be in the pixel as in CenterNet. $\alpha=0.54$ is set in our network, and it's not carefully selected.

The peak of the Gaussian distribution, also the pixel at the box center, is treated as the positive sample while any other pixel is treated as the negative sample. We use modified focal loss as  \cite{law2018cornernet,DBLP:journals/corr/abs-1904-07850}.

Given the prediction $\hat H$ and localization target $H$, we have

\begin{equation} \label{loc_loss}
\begin{aligned}
L_{loc} = \frac{1}{M}\sum_{xyc}
  \left\{ 
  \begin{aligned}
    & (1-\hat H_{ijc})^{\alpha_f} \text{log}(\hat H_{ijc}) ~~~~~~~~~\text{if } H_{ijc}=1 \\
    & (1-H_{ijc})^{\beta_f} \hat H_{ijc}^{\alpha_f} \text{log}(1-\hat H_{ijc}) ~~~~~~~\text{else} \\
  \end{aligned}
  \right.
\end{aligned}
\end{equation}

\noindent where $\alpha_f$ and $\beta_f$ are hyper-parameters in focal loss \cite{lin2017focal} and its modified version \cite{law2018cornernet,DBLP:journals/corr/abs-1904-07850}, respectively. $M$ stands for the number of annotated boxes. We set $\alpha_f=2$ and $\beta_f=4$.

\begin{figure}[!t]
  \centering
  \includegraphics[width=.95\columnwidth]{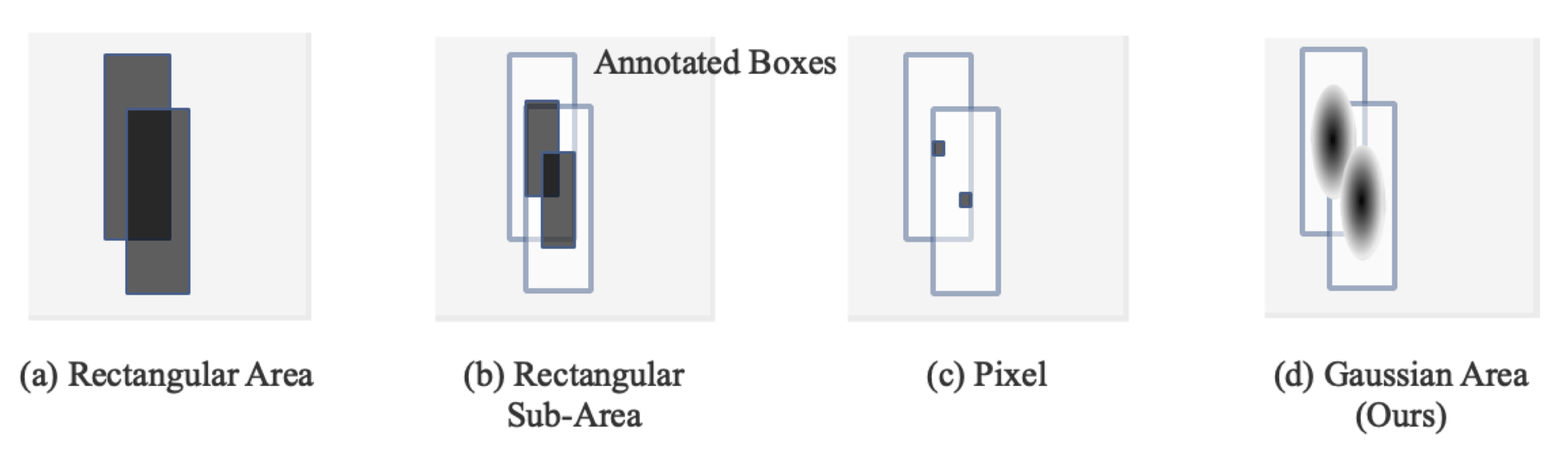}
\caption{Different strategies for defining training samples. Each pixel in the dark area corresponds to a training sample. In (d), the darker the color, the greater the sample weight.}
  \label{reg}
\end{figure}

\paragraph{Size Regression.} Given $m$-th annotated box on the feature-map scale, another Gaussian kernel is adopted to produce $S_m \in R^{1\times \frac{H}{r} \times \frac{W}{r}}$. The kernel size is controlled by $\beta$ as mentioned above. Note that we can use the same kernel to save computation when $\alpha$ and $\beta$ are the same. The non-zero part in $S_m$ is named Gaussian-area $A_m$, as shown in Figure \ref{reg}. Since $A_m$ is always inside the $m$-box, it is also named sub-area in the rest of the paper.

Each pixel in the sub-area is treated as a regression sample. Given pixel $(i,j)$ in the area $A_m$ and output stride $r$, the regression target is defined as the distances from $(ir,jr)$ to four sides of $m$-th box, represented as a $4$-dim vector $(w_l, h_t, w_r, h_b)^m_{ij}$. The predicted box at $(i,j)$ can be represented as:

\begin{equation} \label{regression_targets}
\begin{aligned}
 \hat x_1 = i r - \hat w_l s, ~~ \hat y_1 = j r - \hat h_t s, \\
 \hat x_2 = i r + \hat w_r s, ~~ \hat y_2 = j r + \hat h_b s.
\end{aligned}
\end{equation}

\noindent where $s$ is a fixed scalar used to enlarge the predicted results for easier optimization. $s=16$ is set in our experiments. Note that the predicted box $(\hat x_1, \hat y_1,\hat x_2,\hat y_2)$ is on image scale rather than feature-map scale.

If a pixel is not contained in any sub-areas, it will be ignored during training. If a pixel is contained in multiple sub-areas --- an ambiguous sample, its training target is set to the object with the smaller area.

Given the prediction $\hat S$ and regression target $S$, we gather training targets $S'\in  R^{N_{reg}\times 4}$ from $S$ and corresponding prediction results $\hat S'\in R^{N_{reg}\times 4}$ from $\hat S$, where $N_{reg}$ stands for the number of regression samples. For all these samples, we decode the predicted boxes and corresponding annotated boxes of samples as in (\ref{regression_targets}), and we use GIoU \cite{rezatofighi2019generalized} for loss calculation.

\begin{equation} \label{regression_loss}
\begin{aligned}
  L_{reg} = \frac{1}{N_{reg}}\sum_{(i,j)\in A_m}
    \text{GIoU}(\hat B_{ij}, B_m)\times W_{ij}
\end{aligned}
\end{equation}

\noindent where $\hat B_{ij}$ stands for the decoded box $(\hat x_1, \hat y_1,  \hat x_2,  \hat y_2)_{ij}$ and $B_m = (x_1,y_1,x_2,y_2)_m$ is the corresponding $m$-th annotated box on image scale. $W_{ij}$ is the sample weight, which is used to balance the loss contributed by each sample.

Due to the large scale variance of objects, large objects may produce thousands of samples, whereas small objects may only produce a few. After normalizing the loss contributed by all samples, the losses contributed by small objects are even negligible, which will harm the detection performance on small objects. Therefore, sample weight $W_{ij}$ plays an important role in balancing losses. Suppose $(i,j)$ is inside the sub-area $A_m$ of $m$-th annotated box, we have:

\begin{equation} \label{regression_targets_weight}
\begin{aligned}
  W_{ij} = 
  \left\{ 
  \begin{aligned}
    & \text{log}(a_m)\times \frac{\text{G}_m(i,j)}{\sum_{(x,y)\in A_m}\text{G}_m(x,y)}  & (i,j)\in A_m \\
    & 0 & (i,j) \notin A~~~ \\
  \end{aligned}
  \right.
\end{aligned}
\end{equation}

\noindent where G$_m(i, j)$ is the Gaussian probabilities at $(i, j)$ and $a_m$ is the area of the $m$-th box. This scheme can make good use of more annotation information contained in large objects and preserve that of small objects. It also can emphasize these samples near the object center, reducing the effect of ambiguous and low-quality samples, which will be discussed in our ablation study.

\paragraph{Total Loss}

The total loss $L$ is composed of localization loss $L_{loc}$ and regression loss $L_{reg}$, weighted by two scalar. Specifically, $L=w_{loc}L_{loc}+w_{reg}L_{reg}$, where $w_{loc}=1.0$ and $w_{reg}=5.0$ in our setting.

\subsection{Overall Design}

The architecture of TTFNet is shown in Figure \ref{archi}. We use ResNet and DarkNet \cite{redmon2018yolov3} as the backbone in our experiments. The features extracted by the backbone are up-sampled to 1/4 resolution of the original image, which is implemented by Modulated Deformable Convolution (MDCN) \cite{zhu2019deformable} and up-sample layer. MDCN layers are followed by Batch Normalization (BN) \cite{DBLP:conf/icml/IoffeS15} and ReLU. 

The up-sampled features then separately go through two heads for different goals. The localization head produces high activations on those positions near the object center while the regression head directly predicts the distance from those positions to the four sides of the box. Since the object center corresponds to the local maximum at the feature map, we can safely suppress non-maximum values with the help of 2D max-pooling as in \cite{law2018cornernet,DBLP:journals/corr/abs-1904-07850}. Then we use the positions of local maximums to gather regression results. Finally, the detection results can be attained.

Our approach makes efficient use of annotation information contained in large and medium-sized objects, but the promotion is limited for small objects that contain little information. In order to improve the detection performance on small objects in a short training schedule, we add the shortcut connections to introduce high-resolution but low-level features. The shortcut connections introduce the features from stage 2, 3, and 4 of the backbone, and each connection is implemented by $3\times 3$ convolution layers. The number of the layers are set to 3, 2, and 1 for stage 2, 3, and 4, and ReLU follows each layer except for the last one in the shortcut connnection.

\section{Experiments}

\subsection{Experimental Setting}

\paragraph{Dataset.} Our experiments are based on the challenging MS COCO 2017 benchmark. We use the Train split (115K images) for training and report the performances on Val split (5K images).

\paragraph{Training Details.} We use ResNet and DarkNet as the backbone for experiments. We resize the images to $512\times 512$ and do not keep the aspect ratio. Only the random flip is used for data augmentation in training unless the long training schedule is adopted. We use unfrozen BN but freeze all parameters of stem and stage 1 in the backbone. For ResNet, the initial learning rate is 0.016, and the mini-batch size is 128. For DarkNet, the initial learning rate is 0.015, and the mini-batch size is 96. The learning rate is reduced by a factor of 10 at epoch 18 and 22, respectively. Our network is trained with SGD for 24 epochs. For the super-fast version, the training schedule is halved. For the long-training version, the training schedule is increased by five times(i.e., 120-epochs training in total). Weight decay and momentum are set as 0.0004 and 0.9, respectively. For bias parameters in the network, their weight decay is set to 0, and their learning rate is doubled. Warm-up is applied for the first 500 steps. We initialize our backbone networks with the weights pre-trained on ImageNet \cite{deng2009imagenet} unless specified. Our experiments are based on open source detection toolbox mmdetection \cite{DBLP:journals/corr/abs-1906-07155} with 8 GTX 1080Ti.

\subsection{Ablation Study}
We use the super-fast TTNet-53 in our ablation study. The AP is reported on COCO 5k-val, and the inference speed is measured on the converged model with 1 GTX 1080Ti.

\begin{table}[!tbh] 
\centering
\resizebox{.95\columnwidth}{!}{
\begin{tabular}{l|c|c|c|c|c|c}
\hline
      + Norm & & \checkmark & \checkmark & \checkmark & \checkmark & \checkmark\\ 
      + Sqrt & & & \checkmark & & \checkmark & \\ 
      + Log & & & & \checkmark & & \checkmark \\ 
      + Gaussian & & & & & \checkmark & \checkmark \\ \hline
      AP &  27.2 & 31.7 & 31.2 & 32.0 & 31.6 & 32.9\\
\end{tabular}}
\caption{Different settings of regression weights $W_{ij}$. Norm stands for equally treating $n$ batches of samples produced by $n$ objects, Sqrt and Log stand for multiplying the sample weight by square root or logarithm of the box area, and Gaussian stands for using Gaussian probabilities in sample weights. $\beta$ is set to 0.54 in the experiments.}
\label{wh_weight}
\end{table}

\paragraph{Regression Weight $W_{ij}$.} Each annotated box produces multiple training samples in training, so how to balance the losses produced by samples becomes a problem. Treating all samples equally will lead to poor precision, as shown in Table \ref{wh_weight}. The poor result is caused by the number of samples produced by large objects is hundreds of times larger than that of small objects, which makes the losses contributed by small objects almost negligible. 

Another straightforward method is to normalize the losses produced by each annotated box. Namely, all these samples have same weight $\frac{1}{n_m}$ if $m$-th annotated box produces $n_m$ samples. Still, it leads to sub-optimal results since it loses the chance to utilize the more information contained in large boxes.

To address this problem, we adopt the logarithm of the box area together with the normalized Gaussian probability as the sample weight. The results are listed in Table \ref{wh_weight}, which show that our strategy can greatly handle the issues above. Note that introducing Gaussian probability in the weight can bring other benefits, which will be discussed next.

\begin{table}[!t]
\centering
\resizebox{.95\columnwidth}{!}{
\begin{tabular}{c|c|c|c|c|c|c|c|c}
\hline
& $\beta$ & 0.01 & 0.1 & 0.2 & 0.3 & 0.5 & 0.7 & 0.9 \\
\hline
\multirow{2}{*}{Agnostic} & AP & 27.2 & 31.0 & 32.1 & \textbf{32.5} & 32.2 & 30.9 & 29.5 \\\cline{2-9}
& Ratio & 0.40 & 1.14 & 2.17 &\textbf{3.65} & 6.81 & 11.30 & 17.27 \\
\hline
\multirow{2}{*}{Aware} & AP & 26.7 & 30.9 & 32.1 & 32.0 & 30.9 & 30.1 & 22.6 \\ \cline{2-9}
& Ratio & 0.05 & 0.09 & 0.22 & 0.42 & 1.18 & 2.49 & 4.34
\end{tabular}}
\caption{Results of changing $\beta$ from 0.01 to 0.9 and adopting class-aware regression. Note that the sub-area here is a rectangle, and Norm+Log is used as the sample weights in the experiments. Ratio stands for the relative number of ambiguous samples in the training set.}
\label{wh_beta}
\end{table}

\begin{table}[!t]
\centering
\resizebox{.95\columnwidth}{!}{
\begin{tabular}{c|c|c|c|c}
\hline
     $\beta$ & w/ Gaussian & w/ Aspect Ratio & AP & Ratio \% \\
\hline
0.3 & & & 32.2 & 3.66 \\
0.3 & & \checkmark & 32.5 & 3.65 \\
0.54 & & \checkmark & 32.0 & 8.01 \\
0.54 & \checkmark & & 32.7 & 7.57 \\
0.54 &\checkmark & \checkmark & \textbf{32.9} & \textbf{7.13} \\
\end{tabular}}
\caption{Results of different kernels for producing samples. Gaussian stands for producing the regression samples using Gaussian kernel, and Aspect Ratio stands for considering the aspect ratio of the box in the kernel. $\beta=0.54$ is set to be consistent with $\alpha=0.54$, which allows us to share the same Gaussian kernel for both localization and regression. Ratio stands for the relative number of ambiguous samples in the training set.}
\label{comp_wh}
\end{table}

\begin{table}[!t]
\centering
\begin{tabular}{c|c|c|c|c|c|c}
\hline
Stage 2 & 0 & 1 & 2 & 2 & 3 & 3 \\
Stage 3 & 0 & 1 & 1 & 2 & 2 & 3 \\
Stage 4 & 0 & 1 & 1 & 1 & 1 & 1 \\ \hline
AP & 29.0 & 32.0 & 32.8 & 32.8 & 32.9 & 33.2 \\
FPS & 58.7 & 55.5 & 54.8 & 54.6 & 54.4 & 54.3  \\
\end{tabular}
\caption{Speed-Accuracy tradeoffs when using different settings in shortcut connection.}
\label{shortcut}
\end{table}

\begin{table}[!tbh]
\centering
\resizebox{.95\columnwidth}{!}{
\begin{tabular}{c|c|c|c|c|c|c}
\hline
LR & \multicolumn{2}{|c}{6e-3} & \multicolumn{2}{|c}{1.2e-2} & \multicolumn{2}{|c}{1.8e-2} \\
\hline
Schedule & 1x & 2x & 1x & 2x & 1x & 2x \\
\hline
$\beta_1=0.01$ & \textbf{29.9} & \textbf{33.4} & 29.2 & 33.1 & 2.9 & 6.0 \\
$\beta_2=0.03$ & \textbf{30.1} & \textbf{33.5} & 29.4 & 33.1 & 7.4 & 20.2 \\
$\beta_3=0.1$ & \textbf{30.9} & 33.7 & 30.0 & \textbf{33.8} & 28.1 & 32.7 \\
$\beta_4=0.2$ & 31.0 & 33.8 & \textbf{31.8} & \textbf{34.4} & 30.6 & 34.0 \\
$\beta_5=0.4$ & 31.8 & 34.3 & \textbf{32.6} & 35.0 & 32.2 & \textbf{35.2} \\
$\beta_6=0.54$ & 31.9 & 34.1 & 32.5 & 35.0 & \textbf{32.6} & \textbf{35.3}
\end{tabular}}
\caption{Results of different kernel size $\beta$. $\alpha=0.54$ and Gaussian kernel is used to produce regression samples. 1x stands for 12-epochs training and 2x stands for 24-epochs training.}
\label{relation}
\end{table}

\begin{table}[!t]
\centering
\begin{tabular}{c|c|c|c|c}
\hline
\multirow{2}{*}{Schedule} & \multirow{2}{*}{w/Pre-Train}  & \multicolumn{3}{|c}{Backbone} \\ \cline{3-5}
& & R18 & R34 & D53 \\
\hline
2x & \checkmark & 28.1 & 31.3 & 35.1 \\
\hline
10x & & 30.3 & 33.3 & 36.2 \\
10x &\checkmark & 31.8 & 35.3 & 39.3 \\
\end{tabular}
\caption{AP after adopting long training schedule. We use data augmentation to prevent overfitting.}
\label{scratch}
\end{table}

\begin{table*}[tbp]
\centering
\resizebox{1.7\columnwidth}!{
\begin{tabular}{lllllllllll}
\Xhline{2\arrayrulewidth}
Method             & Backbone       & Size    & FPS          & TT(h) & AP   & AP$_{50}$ & AP$_{75}$ & AP$_S$  & AP$_M$  & AP$_L$  \\
\hline
RetinaNet \cite{lin2017focal} *         & R18-FPN  & 1330, 800 & 16.3            &  6.9              & 30.9 & 49.6 & 32.7 & 15.8 & 33.9 & 41.9 \\
RetinaNet \cite{lin2017focal} *         & R34-FPN  & 1330, 800 & 15.0            &  8.3              & 34.7 & 54.0 & 37.3 & 18.2 & 38.6 & 45.9 \\
RetinaNet \cite{lin2017focal}          & R50-FPN  & 1330, 800 & 12.0            & 11.0           & 35.8 & 55.4 & 38.2 & 19.5 & 39.7 & 46.6 \\
FCOS \cite{DBLP:journals/corr/abs-1904-01355} *          & R18-FPN  & 1330, 800 & 20.8            & 5.0            & 26.9 & 43.2 & 27.9 & 13.9 & 28.9 & 36.0 \\
FCOS \cite{DBLP:journals/corr/abs-1904-01355} *              & R34-FPN  & 1330, 800 & 16.3            & 6.0            & 32.2 & 49.5 & 34.0 & 17.2 & 35.2 & 42.1 \\
FCOS \cite{DBLP:journals/corr/abs-1904-01355}               & R50-FPN  & 1330, 800 & 15.0            & 7.8            & 36.6 & 55.8 & 38.9 & 20.8 & 40.3 & 48.0 \\
\hline
SSD \cite{liu2016ssd}               & VGG16    & 300, 300  & 44.0            & 21.4           & 25.7 & 43.9 & 26.2 & 6.9  & 27.7 & 42.6 \\
SSD \cite{liu2016ssd}               & VGG16    & 512, 512  & 28.4            & 36.1           & 29.3 & 49.2 & 30.8 & 11.8 & 34.1 & 44.7 \\
YOLOv3 \cite{redmon2018yolov3}            & D53      & 320, 320  & 55.7            & 26.4           & 28.2 & -    & -    & -    & -    & -    \\
YOLOv3 \cite{redmon2018yolov3}             & D53      & 416, 416  & 46.1            & 31.6           & 31.0 & -    & -    & -    & -    & -    \\
YOLOv3 \cite{redmon2018yolov3}             & D53      & 608, 608  & 30.3            & 66.7           & 33.0 & 57.9 & 34.4 & 18.3 & 25.4 & 41.9 \\
CenterNet \cite{DBLP:journals/corr/abs-1904-07850}         & R18      & 512, 512  & 128.5           & 26.9           & 28.1 & 44.9 & 29.6 & -    & -    & -    \\
CenterNet \cite{DBLP:journals/corr/abs-1904-07850}         & R101     & 512, 512  & 44.7            & 49.3           & 34.6 & 53.0 & 36.9 & -    & -    & -   \\
CenterNet \cite{DBLP:journals/corr/abs-1904-07850}         & DLA34     & 512, 512  & 55.0            & 46.8           & 37.4 & 55.1 & 40.8 & -    & -    & -   \\
\hline
TTFNet (\textit{fast})             & R18      & 512, 512  & 112.2           & \textbf{1.8}            & 25.9 & 41.3 & 27.9 & 10.7 & 27.1 & 38.6 \\
TTFNet             & R18      & 512, 512  & 112.3           & \textbf{3.6}            & 28.1 & 43.8 & 30.2 & 11.8 & 29.5 & 41.5  \\
TTFNet             & R34      & 512, 512  & 86.6            & \textbf{4.1}            & 31.3 & 48.3 & 33.6 & 13.5 & 34.0 & 45.7 \\
TTFNet (\textit{fast})       & D53      & 512, 512  & 54.8            & \textbf{3.1}            & 32.9 & 50.2 & 35.9 & 15.3 & 36.1 & 45.2 \\
TTFNet             & D53      & 512, 512  & 54.4            & \textbf{6.1}            & 35.1 & 52.5 & 37.8 & 17.0 & 38.5 & 49.5 \\
TTFNet (\textit{10x})            & D53      & 512, 512  & 57.0            & 30.6            & \textbf{39.3} & 56.8 & 42.5 & 20.6 & 43.3 & 54.3 \\
\hline
\end{tabular}}
\caption{TTFNet vs. other state-of-the-art one-stage detectors. TT stands for training time. * indicates that the result is not presented in the original paper. \textit{fast} stands for the super-fast version and \textit{10x} stands for the long-training version. All the training time is measured on 8 GTX 1080Ti, and all the inference speed is measured using converged models on 1 GTX 1080Ti.}
\label{compare}
\end{table*}

\paragraph{Benefits of Gaussian Design in Regression.} We introduce the Gaussian probability in the regression weight, which can reduce the impact of ambiguous samples and low-quality samples more elegantly and efficiently. The ambiguous sample refers to the sample located in the overlapped area, and the low-quality sample refers to the sample that is far away from the object center. 

Specifically, multiple objects are spatially overlapped sometimes, and thus it is hard for anchor-free detectors to decide the regression target in the overlapping area, which is called ambiguity. To alleviate it, previous work either places object of different scale in different level by using FPN \cite{DBLP:journals/corr/abs-1904-01355,DBLP:journals/corr/abs-1904-03797}, or produces just one training sample based on the annotated box \cite{DBLP:journals/corr/abs-1904-07850}, as shown in Figure \ref{reg}. Previous work \cite{DBLP:journals/corr/abs-1904-01355} also has noticed the impact of low-quality sample, and it suppresses low-quality samples by introducing the "center-ness" prediction. However, these solutions can only reduce the ambiguous samples or the low-quality samples. Besides, they have some side effects, such as leading to slow convergence speed or inference speed.

Our Gaussian design can reduce both of the two types of samples without any side effects. It produces a sub-area inside the box, and the relative size of the sub-area is affected by the hyper-parameter $\beta$. Larger $\beta$ utilizes more annotated information but also brings more ambiguous samples and low-quality samples.

Firstly, we use a more mundane form, i.e., rectangle as the sub-area to analyze the relationship between precision and $\beta$. In particular, $\beta=0$ means only the box center is treated as a regression sample as in CenterNet, while $\beta=1$ means all pixels in the rectangle box are treated as regression samples. We train a series of networks with changing $\beta$ from 0.01 to 0.9. As shown in Table \ref{wh_beta}, the AP first rises and then falls as $\beta$ increases. The rise indicates the annotated information near the object center also matters --- the AP when $\beta=0.3$ is much higher than that when $\beta=0.01$. Therefore, the strategy of CenterNet that merely considers the object center is sub-optimal. The decline is caused by the increased ambiguous samples and low-quality samples. To find out the main factor, we conduct experiments with the class-aware regression. Results show that we still meet the obvious accuracy degradation even the class-aware regression has reduced the impact of the ambiguity. It reveals that the main reason of the decline is caused by those low-quality samples.

So, then, we propose the approach that uses Gaussian kernel to produce sub-area for training samples. Our approach not only emphasizes the samples near the object center but also alleviates the ambiguity. As shown in Table \ref{comp_wh}, using the Gaussian sub-area achieves better results than using rectangular sub-area.

\paragraph{Considering the Aspect Ratio of the Box in Gaussian Kernel.} CenterNet adopts the same strategy as CornetNet to produce heat-map without considering the aspect ratio of the box. According to our experiments, considering the ratio can improve precision consistently, as shown in Table \ref{comp_wh}.

\paragraph{Shortcut Connection.} We introduce the shortcut connection for achieving higher precision. The results when using different settings are listed in Table \ref{shortcut}. We choose the combination of 3, 2, 1 for stage 2, 3, 4, and it is not carefully selected.

\paragraph{The Effect of Sample Number on the Learning Rate.} To verify the similarity between the batch size and the number of training samples encoded by the annotated boxes, we conduct experiments by changing $\beta$ and learning rate.

As shown in Table \ref{relation}, we can observe that larger $\beta$ guarantees a larger learning rate and better performance. Besides, the trend is more noticeable when $\beta$ is smaller since there are fewer ambiguous and low-quality samples. In other words, having more samples is like enlarging the batch size, which helps to increase the learning rate further.

\paragraph{Training from Scratch.} The from-scratch-training usually requires a longer training schedule. We set the total training epochs to 120 here. As shown in Table \ref{scratch}, from-scratch-training models can achieve performances comparable to those having a pre-trained backbone. Moreover, much better performance can be achieved when using a long training schedule, but it takes much longer training time.

\begin{table}[!t]
\centering
\resizebox{.95\columnwidth}{!}{
\begin{tabular}{c|c|c|c|c}
\hline
Method & Backbone & Schedule & w/Augmentation & AP \\
\hline
CenterNet & R18 & 2x & \checkmark & 20.0 \\
CenterNet & R18 & 2x & & 20.8 \\
TTFNet & R18 & 2x & & 28.1 \\
\hline
CenterNet & R18 & 11.67x & \checkmark & 28.1 \\
TTFNet & R18 & 10x & \checkmark & 31.8 \\
\hline
CenterNet & DLA34 & 2x & \checkmark & 26.2 \\
CenterNet & DLA34 & 2x & & 31.6 \\
TTFNet & DLA34 & 2x & & 34.9 \\
\hline
CenterNet & DLA34 & 11.67x & \checkmark & 37.4 \\
TTFNet & DLA34 & 10x & \checkmark & 38.2 \\
\end{tabular}}
\caption{TTFNet vs. CenterNet.}
\label{ctnet-ttfnet}
\end{table}

\subsection{Compared with State-of-the-Arts Detectors}

Our TTFNet adopts ResNet-18/34 and DarkNet-53 as the backbone, and they are marked as TTFNet-18/34/53. As shown in Table \ref{compare}, our network can be more than seven times faster than other real-time detectors in training time while achieving state-of-the-art results with real-time inference speed. Compared with SSD300, our super-fast TTFNet-18 can achieve slightly higher precision, but our training time is ten times less, and the inference is more than two times faster. As for YOLOv3, our TTFNet-53 can achieve 2 points higher precision in just one-tenth training time, and it is almost two times faster than YOLOv3 in inference. The super-fast TTFNet-53 can reach the precision of YOLOv3 in just one-twentieth training time.

As for the recently proposed anchor-free detector, our TTFNet shows great advantages. FCOS can achieve high precision without requiring long training time, but its slow inference speed will limit its mobile application. We list the performance of adopting lighter backbone such as ResNet18/34 in advanced RetinaNet and FCOS. Unfortunately, they can not achieve comparable performance due to the heavy head design. As for the real-time detector CenterNet, it has very fast inference speed and high precision, but it requires long training time. Our TTFNet only needs one-seventh training time compared with CenterNet, and it is superior in balancing training time, inference speed, and accuracy.

\paragraph{More Comparisons with CenterNet.} CenterNet achieves 37.4 AP after being trained for 140 epochs when using DLA34\shortcite{DBLP:conf/cvpr/YuWSD18} as the backbone. We notice that CenterNet uses specially customized up-sampling layers for DLA34. For comparison between CenterNet and TTFNet when using DLA34, we replace the up-sampling layers in TTFNet with the ones in CenterNet, and therefore our changes in network structures cannot be applied. We use the same training hyper-parameters as TTFNet-53. The results in Table \ref{ctnet-ttfnet} show that our approach can bring significant improvements.

\section{Conclusion}

We empirically show that more training samples help enlarge the learning rate and propose the novel method of using the Gaussian kernel for training. It is an elegant and efficient solution for balancing training time, inference speed, and accuracy, which can provide more potentials and possibilities for training-time-sensitive tasks\cite{DBLP:conf/iclr/ZophL17,DBLP:conf/cvpr/ZophVSL18,ghiasi2019fpn,DBLP:journals/corr/abs-1906-04423,gao2019nddr}.

\section*{Acknowledgments}

This work was supported in part by The National Key Research and Development Program of China (Grant Nos: 2018AAA0101400), in part by The National Nature Science Foundation of China (Grant Nos: 61936006, 61973271).

\small \bibliography{3625-aaai.bib}

\begin{thebibliography}{}

\bibitem[\protect\citeauthoryear{Chen \bgroup et al\mbox.\egroup
  }{2019}]{DBLP:journals/corr/abs-1906-07155}
Chen, K.; Wang, J.; Pang, J.; Cao, Y.; Xiong, Y.; Li, X.; Sun, S.; Feng, W.;
  Liu, Z.; Xu, J.; Zhang, Z.; Cheng, D.; Zhu, C.; Cheng, T.; Zhao, Q.; Li, B.;
  Lu, X.; Zhu, R.; Wu, Y.; Dai, J.; Wang, J.; Shi, J.; Ouyang, W.; Loy, C.~C.;
  and Lin, D.
\newblock 2019.
\newblock Mmdetection: Open mmlab detection toolbox and benchmark.
\newblock {\em CoRR} abs/1906.07155.

\bibitem[\protect\citeauthoryear{Deng \bgroup et al\mbox.\egroup
  }{2009}]{deng2009imagenet}
Deng, J.; Dong, W.; Socher, R.; Li, L.-J.; Li, K.; and Fei-Fei, L.
\newblock 2009.
\newblock Imagenet: A large-scale hierarchical image database.
\newblock In {\em 2009 IEEE conference on computer vision and pattern
  recognition},  248--255.
\newblock Ieee.

\bibitem[\protect\citeauthoryear{Duan \bgroup et al\mbox.\egroup
  }{2019}]{DBLP:journals/corr/abs-1904-08189}
Duan, K.; Bai, S.; Xie, L.; Qi, H.; Huang, Q.; and Tian, Q.
\newblock 2019.
\newblock Centernet: Keypoint triplets for object detection.
\newblock {\em CoRR} abs/1904.08189.

\bibitem[\protect\citeauthoryear{Gao \bgroup et al\mbox.\egroup
  }{2019}]{gao2019nddr}
Gao, Y.; Ma, J.; Zhao, M.; Liu, W.; and Yuille, A.~L.
\newblock 2019.
\newblock {NDDR}-{CNN}: Layerwise feature fusing in multi-task cnns by neural
  discriminative dimensionality reduction.
\newblock In {\em IEEE International Conference on Computer Vision and Pattern
  Recognition (CVPR)}.

\bibitem[\protect\citeauthoryear{Ghiasi, Lin, and Le}{2019}]{ghiasi2019fpn}
Ghiasi, G.; Lin, T.-Y.; and Le, Q.~V.
\newblock 2019.
\newblock Nas-fpn: Learning scalable feature pyramid architecture for object
  detection.
\newblock In {\em Proceedings of the IEEE Conference on Computer Vision and
  Pattern Recognition},  7036--7045.

\bibitem[\protect\citeauthoryear{Goyal \bgroup et al\mbox.\egroup
  }{2017}]{DBLP:journals/corr/GoyalDGNWKTJH17}
Goyal, P.; Doll{\'{a}}r, P.; Girshick, R.~B.; Noordhuis, P.; Wesolowski, L.;
  Kyrola, A.; Tulloch, A.; Jia, Y.; and He, K.
\newblock 2017.
\newblock Accurate, large minibatch {SGD:} training imagenet in 1 hour.
\newblock {\em CoRR} abs/1706.02677.

\bibitem[\protect\citeauthoryear{Huang \bgroup et al\mbox.\egroup
  }{2015}]{DBLP:journals/corr/HuangYDY15}
Huang, L.; Yang, Y.; Deng, Y.; and Yu, Y.
\newblock 2015.
\newblock Densebox: Unifying landmark localization with end to end object
  detection.
\newblock {\em CoRR} abs/1509.04874.

\bibitem[\protect\citeauthoryear{Ioffe and
  Szegedy}{2015}]{DBLP:conf/icml/IoffeS15}
Ioffe, S., and Szegedy, C.
\newblock 2015.
\newblock Batch normalization: Accelerating deep network training by reducing
  internal covariate shift.
\newblock In {\em Proceedings of the 32nd International Conference on Machine
  Learning, {ICML} 2015, Lille, France, 6-11 July 2015},  448--456.

\bibitem[\protect\citeauthoryear{Kong \bgroup et al\mbox.\egroup
  }{2019}]{DBLP:journals/corr/abs-1904-03797}
Kong, T.; Sun, F.; Liu, H.; Jiang, Y.; and Shi, J.
\newblock 2019.
\newblock Foveabox: Beyond anchor-based object detector.
\newblock {\em CoRR} abs/1904.03797.

\bibitem[\protect\citeauthoryear{Law and Deng}{2018}]{law2018cornernet}
Law, H., and Deng, J.
\newblock 2018.
\newblock Cornernet: Detecting objects as paired keypoints.
\newblock In {\em Proceedings of the European Conference on Computer Vision
  (ECCV)},  734--750.

\bibitem[\protect\citeauthoryear{Lin \bgroup et al\mbox.\egroup
  }{2014}]{lin2014microsoft}
Lin, T.-Y.; Maire, M.; Belongie, S.; Hays, J.; Perona, P.; Ramanan, D.;
  Doll{\'a}r, P.; and Zitnick, C.~L.
\newblock 2014.
\newblock Microsoft coco: Common objects in context.
\newblock In {\em European conference on computer vision},  740--755.
\newblock Springer.

\bibitem[\protect\citeauthoryear{Lin \bgroup et al\mbox.\egroup
  }{2017a}]{lin2017feature}
Lin, T.-Y.; Doll{\'a}r, P.; Girshick, R.; He, K.; Hariharan, B.; and Belongie,
  S.
\newblock 2017a.
\newblock Feature pyramid networks for object detection.
\newblock In {\em Proceedings of the IEEE conference on computer vision and
  pattern recognition},  2117--2125.

\bibitem[\protect\citeauthoryear{Lin \bgroup et al\mbox.\egroup
  }{2017b}]{lin2017focal}
Lin, T.-Y.; Goyal, P.; Girshick, R.; He, K.; and Doll{\'a}r, P.
\newblock 2017b.
\newblock Focal loss for dense object detection.
\newblock In {\em Proceedings of the IEEE international conference on computer
  vision},  2980--2988.

\bibitem[\protect\citeauthoryear{Liu \bgroup et al\mbox.\egroup
  }{2016}]{liu2016ssd}
Liu, W.; Anguelov, D.; Erhan, D.; Szegedy, C.; Reed, S.; Fu, C.-Y.; and Berg,
  A.~C.
\newblock 2016.
\newblock Ssd: Single shot multibox detector.
\newblock In {\em European conference on computer vision},  21--37.
\newblock Springer.

\bibitem[\protect\citeauthoryear{Redmon and Farhadi}{2017}]{redmon2017yolo9000}
Redmon, J., and Farhadi, A.
\newblock 2017.
\newblock Yolo9000: better, faster, stronger.
\newblock In {\em Proceedings of the IEEE conference on computer vision and
  pattern recognition},  7263--7271.

\bibitem[\protect\citeauthoryear{Redmon and Farhadi}{2018}]{redmon2018yolov3}
Redmon, J., and Farhadi, A.
\newblock 2018.
\newblock Yolov3: An incremental improvement.
\newblock {\em arXiv preprint arXiv:1804.02767}.

\bibitem[\protect\citeauthoryear{Redmon \bgroup et al\mbox.\egroup
  }{2016}]{redmon2016you}
Redmon, J.; Divvala, S.; Girshick, R.; and Farhadi, A.
\newblock 2016.
\newblock You only look once: Unified, real-time object detection.
\newblock In {\em Proceedings of the IEEE conference on computer vision and
  pattern recognition},  779--788.

\bibitem[\protect\citeauthoryear{Ren \bgroup et al\mbox.\egroup
  }{2015}]{ren2015faster}
Ren, S.; He, K.; Girshick, R.; and Sun, J.
\newblock 2015.
\newblock Faster r-cnn: Towards real-time object detection with region proposal
  networks.
\newblock In {\em Advances in neural information processing systems},  91--99.

\bibitem[\protect\citeauthoryear{Rezatofighi \bgroup et al\mbox.\egroup
  }{2019}]{rezatofighi2019generalized}
Rezatofighi, H.; Tsoi, N.; Gwak, J.; Sadeghian, A.; Reid, I.; and Savarese, S.
\newblock 2019.
\newblock Generalized intersection over union: A metric and a loss for bounding
  box regression.
\newblock In {\em Proceedings of the IEEE Conference on Computer Vision and
  Pattern Recognition},  658--666.

\bibitem[\protect\citeauthoryear{Tian \bgroup et al\mbox.\egroup
  }{2019}]{DBLP:journals/corr/abs-1904-01355}
Tian, Z.; Shen, C.; Chen, H.; and He, T.
\newblock 2019.
\newblock {FCOS:} fully convolutional one-stage object detection.
\newblock {\em CoRR} abs/1904.01355.

\bibitem[\protect\citeauthoryear{Wang \bgroup et al\mbox.\egroup
  }{2019a}]{wang2019region}
Wang, J.; Chen, K.; Yang, S.; Loy, C.~C.; and Lin, D.
\newblock 2019a.
\newblock Region proposal by guided anchoring.
\newblock In {\em Proceedings of the IEEE Conference on Computer Vision and
  Pattern Recognition},  2965--2974.

\bibitem[\protect\citeauthoryear{Wang \bgroup et al\mbox.\egroup
  }{2019b}]{DBLP:journals/corr/abs-1906-04423}
Wang, N.; Gao, Y.; Chen, H.; Wang, P.; Tian, Z.; and Shen, C.
\newblock 2019b.
\newblock {NAS-FCOS:} fast neural architecture search for object detection.
\newblock {\em CoRR} abs/1906.04423.

\bibitem[\protect\citeauthoryear{Yu \bgroup et al\mbox.\egroup
  }{2016}]{yu2016unitbox}
Yu, J.; Jiang, Y.; Wang, Z.; Cao, Z.; and Huang, T.
\newblock 2016.
\newblock Unitbox: An advanced object detection network.
\newblock In {\em Proceedings of the 24th ACM international conference on
  Multimedia},  516--520.
\newblock ACM.

\bibitem[\protect\citeauthoryear{Yu \bgroup et al\mbox.\egroup
  }{2018}]{DBLP:conf/cvpr/YuWSD18}
Yu, F.; Wang, D.; Shelhamer, E.; and Darrell, T.
\newblock 2018.
\newblock Deep layer aggregation.
\newblock In {\em 2018 {IEEE} Conference on Computer Vision and Pattern
  Recognition, {CVPR} 2018, Salt Lake City, UT, USA, June 18-22, 2018},
  2403--2412.
\newblock {IEEE} Computer Society.

\bibitem[\protect\citeauthoryear{Zhou, Wang, and
  Kr{\"{a}}henb{\"{u}}hl}{2019}]{DBLP:journals/corr/abs-1904-07850}
Zhou, X.; Wang, D.; and Kr{\"{a}}henb{\"{u}}hl, P.
\newblock 2019.
\newblock Objects as points.
\newblock {\em CoRR} abs/1904.07850.

\bibitem[\protect\citeauthoryear{Zhu \bgroup et al\mbox.\egroup
  }{2019}]{zhu2019deformable}
Zhu, X.; Hu, H.; Lin, S.; and Dai, J.
\newblock 2019.
\newblock Deformable convnets v2: More deformable, better results.
\newblock In {\em Proceedings of the IEEE Conference on Computer Vision and
  Pattern Recognition},  9308--9316.

\bibitem[\protect\citeauthoryear{Zhu, He, and
  Savvides}{2019}]{DBLP:journals/corr/abs-1903-00621}
Zhu, C.; He, Y.; and Savvides, M.
\newblock 2019.
\newblock Feature selective anchor-free module for single-shot object
  detection.
\newblock {\em CoRR} abs/1903.00621.

\bibitem[\protect\citeauthoryear{Zoph and Le}{2017}]{DBLP:conf/iclr/ZophL17}
Zoph, B., and Le, Q.~V.
\newblock 2017.
\newblock Neural architecture search with reinforcement learning.
\newblock In {\em 5th International Conference on Learning Representations,
  {ICLR} 2017, Toulon, France, April 24-26, 2017, Conference Track
  Proceedings}.

\bibitem[\protect\citeauthoryear{Zoph \bgroup et al\mbox.\egroup
  }{2018}]{DBLP:conf/cvpr/ZophVSL18}
Zoph, B.; Vasudevan, V.; Shlens, J.; and Le, Q.~V.
\newblock 2018.
\newblock Learning transferable architectures for scalable image recognition.
\newblock In {\em 2018 {IEEE} Conference on Computer Vision and Pattern
  Recognition, {CVPR} 2018, Salt Lake City, UT, USA, June 18-22, 2018},
  8697--8710.

\end{thebibliography}
\bibliographystyle{aaai}
\end{document}